\pdfoutput=1
\documentclass{llncs}
\usepackage{multirow}
\usepackage{caption}
\usepackage{booktabs}
\usepackage{epsfig}
\usepackage{longtable}
\usepackage{rotating}
\usepackage[colorlinks=true,
            linkcolor=red,
            citecolor=blue]{hyperref}

\begin{document}
\frontmatter          

\title{Multi-Scale Convolutional-Stack Aggregation for Robust White Matter Hyperintensities Segmentation}
\author{Hongwei Li\inst{1} \and Jianguo Zhang\inst{3} \and Mark	Muehlau\inst{2} \and Jan Kirschke\inst{2} \and Bjoern~Menze\inst{1}}
\institute{1. Technical University of Munich\\
2. Klinikum rechts der Isar\\
3. University of Dundee, United Kingdom\\
}
\maketitle              
\begin{abstract}
Segmentation of both large and small white matter hyperintensities/lesions in brain MR images is a challenging task which has drawn much attention in recent years.
We propose a multi-scale aggregation model framework to deal with volume-varied lesions. Firstly, we present a specifically-designed network for small lesion segmentation called \emph{Stack-Net}, in which multiple convolutional layers are 'one-by-one' connected, aiming to preserve rich local spatial information of small lesions before the sub-sampling layer. Secondly, we aggregate multi-scale \emph{Stack-Nets} with different receptive fields to learn multi-scale contextual information of both large and small lesions.
Our model is evaluated on recent MICCAI WMH Challenge Dataset and outperforms the state-of-the-art on lesion recall and lesion F1-score under 5-fold cross validation.
It claimed the \textbf{first place} on the hidden test set after independent evaluation by the challenge organizer.
In addition, we further test our pre-trained models on a Multiple Sclerosis lesion dataset with 30 subjects under cross-center evaluation. Results show that the aggregation model is effective in learning multi-scale spatial information.

%

\end{abstract}
\section{Introduction}
White matter hyperintensities (WMH) characterized by bilateral, mostly symmetrical lesions are commonly seen on FLAIR magnetic resonance imaging (MRI) of clinically healthy elderly people; furthermore, they have been repeatedly associated with various neurological and geriatric disorders such as mood problems and cognitive decline \cite{debette2010clinical}. Detection of such lesions on MRI has become a crucial criterion for diagnosis and predicting prognosis in early stage of diseases.

\begin{figure*}[t]
	\begin{center}
		\includegraphics[width=0.95\linewidth,height=0.30\linewidth]{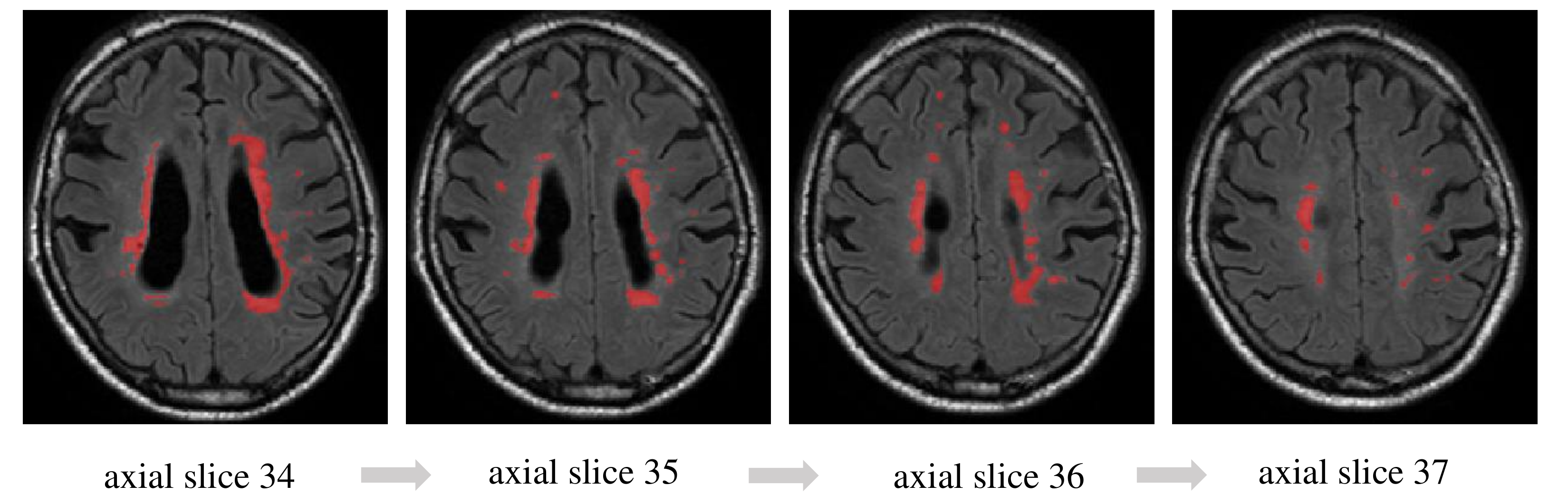}
	\end{center}
	\caption{From left to right: axial slices from 34 to 37 of one case from the MICCAI WMH Challenge public training set, showing the high discontinuity of white matter hyperintensities. The red pixels indicate the WMH annotated by a neuroradiologist. }
	\label{fig:incontinuity} 
\end{figure*}

Different from brain tumor segmentation \cite{menze2015multimodal} in MR images where most of the abnormal regions are large and with spatial continuity, in the task of WMH segmentation, both large and small lesions with high discontinuity are commonly found as shown in Fig. \ref{fig:incontinuity}. 
Generally, small abnormal region contains relatively less contextual information due to the poor spatial continuity.
Furthermore, the feature representation of small lesions tend to be trivial when image features are extracted in a global manner.
One solution to tackle this issue is to use an ensemble model or aggregation model \cite{yu2015multi} to learn different attributes i.e., multi-levels of feature representation from the training data.

Although there exist various computer-aided diagnostic systems for automatic segmentation of white matter hyperintensities \cite{borghesani2013association,moeskops2018evaluation},
the reported results are largely incomparable due to different datasets and evaluation protocols.
The MICCAI WMH Segmentation Challenge 2017 \footnote{\url{http://wmh.isi.uu.nl/}} was the first competition held to compare state-of-the-art algorithms on this task. The winning method \cite{li2018fully} of the challenge employed the modified U-Net \cite{ronneberger2015u} architecture and ensemble models (\emph{U-Net ensembles} in short). Three U-Net models of same architecture were trained with shuffled data and different weight initializations.

In traditional fully convolutional networks \cite{long2015fully}, each convolutional layer is followed by a max-pooling operation which causes the loss of spatial information. In the task of WMH segmentation, this sub-sampling operation can be devastating because small-volume hyperintensities with less than 10 voxels are commonly found. Instead of using single convolutional layer before the sub-sampling layer, we hypothesize that a \emph{convolutional stack} with multiple convolutional layer is able to extract rich local information and it would be more effective by propagating the feature maps learned to the high-resolutional \emph{deconvolutional layer} by skip connections similar to the U-Net approach \cite{ronneberger2015u}.

In this paper, we present a \textit{stacked} architecture of fully convolutional network called \emph{Stack-Net} which aims at
preserving the local spatial information of small lesions and propagating them to deconvolutional layers.
We further aggregate two \emph{Stack-Nets} with different receptive fields to learn multi-scale spatial information from both large and small abnormal regions.
Our method outperforms the state-of-the-art in lesion recall by 4\%  on the MICCAI Challenge Dataset with 60 cases.
In addition, we test our pre-trained models on a private Multiple Sclerosis (MS) lesion dataset with 30 subjects.
Results further demonstrate the effectiveness of the aggregation idea.

\section{Method}

\subsection{Convolutional Stack and Multi-Scale Convolutions}

\subsubsection{Formulation}

Let $f( \cdot )$ represent the nonlinear activation function. The k$^{th}$ output feature map of \emph{l$^{th}$} layer Y$_{lk}$ can be computed as:
\emph{Y$_{lk}$ = f(W$_{lk}^{r}$$\ast$x)}
where the input image is denoted by \emph{x}; the convolutional kernel with fixed size \emph{r$\times$r} related to the \emph{k} feature map is denoted by W$_{lk}^{r}$; the multiplication sign refers to the 2D convolutional operator, which is used to calculate the inner product of the filter model at each location of the input image.
We now generalize this convolution structure from layer to stack.
Let a convolutional stack \emph{S} contain L convolutional layers, the k$^{th}$ output feature map Y$_{Sk}$ of \emph{S} can be computed as
\begin{equation}
	\emph{Y$_{Sk}$ = f(W$_{Lk}^{r}$$\ast$f(W$_{(L-1)k}^{r}$\dots f(W$_{0k}^{r}$$\ast$x))\dots)}
	\end{equation}

\begin{figure*}[t]
	\begin{center}
		\includegraphics[width=1\linewidth,height=0.50\linewidth]{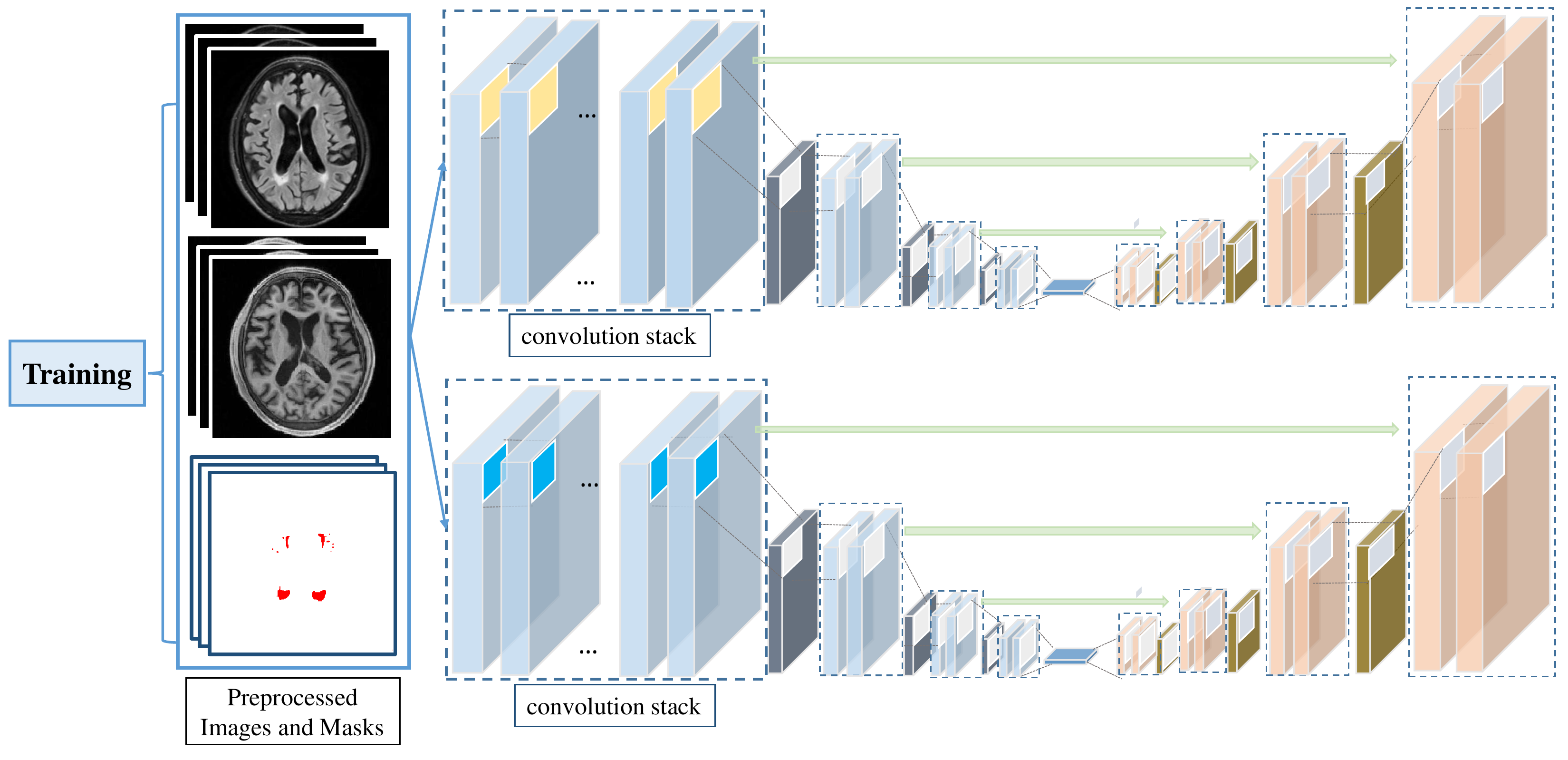}
	\end{center}
	\caption{Overview of the multi-scale convolutional-stack aggregation model.
We replaced the traditional single convolutional-layer with convolutional-stack to extract and preserve local information of small lesions.
The depth of the convolution-stack was flexible and set to 5 in our experiments.
Two convolutional kernels i.e., 3$\times$3 and 5$\times$5 were used in two \emph{Stack-Nets} to learning multi-scale context information.
The detailed parameters setting/architecture was presented in \ref{fig:u_net}.}
	\label{fig:convolution_stack} \vspace{-0.3cm}
\end{figure*}

Obviously, the use of multiple connected convolutional layers to replace single layer would lead to the increase of computational complexity. However, the local spatial information of small lesions could be largely reduced after the first pooling layer (with 2 $\times$ 2 kernel or larger). As a result, we only replaced the first two convolutional layers before sub-sampling layers with convolutional stack as shown in Fig. \ref{fig:convolution_stack}.

We further employed multi-scale convolutional kernels to learn different contextual information from both large and small abnormal regions.
In our task, we aggregate the proposed two \emph{Stack-Nets} with different receptive fields. \emph{Stack-Net} with small receptive field i.e., 3$\times$3 kernel, is expected to learn local spatial information of small-volume hyperintensities while \emph{Stack-Net} with large kernel is designed to learn spatial continuity of large abnormal regions. Two models were trained and optimized independently and were aggregated by using a voting strategy during the testing stage.
Let $P_{i}$ be the 3D segmentation probablity masks predicted by one single model $M_{i}$.
Then the final segmentation probability map of the aggregation of $n$ models is defined as: \emph{$P_{aggr}$} = $\frac{1}{n}$$\sum_{i}P_{i}$.
The threshold for generating binary mask is set to 0.4.

\begin{figure*}[t]
	\begin{center}
		\includegraphics[width=1\linewidth,height=0.45\linewidth]{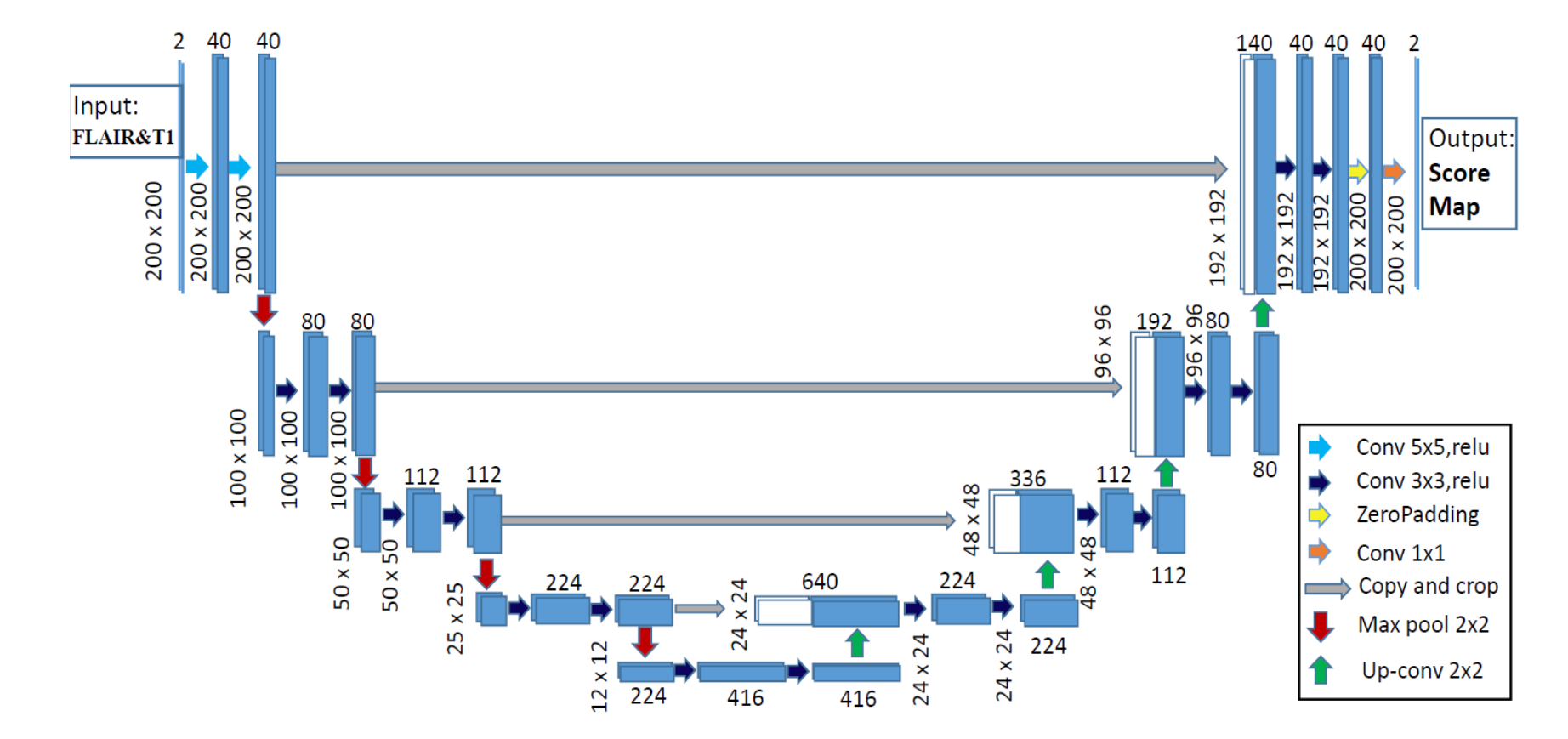}
	\end{center}
	\caption{Detailed parameters setting of the deep networks. The number of stacked layers is 2. }
	\label{fig:u_net} \vspace{-0.3cm}
\end{figure*}

\subsubsection{Architecture}
As shown in Fig. \ref{fig:convolution_stack}, we built two \emph{Stack-Nets} with different convolutional kernels, which takes as input the axial slices (2D) of two modalities from the brain MR scans during both training and testing.
Different from the winning architecture \emph{U-Net Ensembles} \cite{li2018fully} in the MICCAI challenge, 
we replaced the first two convolutional layers by a convolutional stack, with $3\times3$ and $5\times5$ kernel size respectively.
Each convolutional stack is followed by a rectified linear unit (ReLU) and a 2$\times$2 max pooling operation with stride 2 for downsampling. The depth of the convolutional stack was set as \emph{L=5}.
In total the network contains 24 convolutional and de-convolutional layers.
\subsubsection{Training} 
For the data preprocessing, each slice and the corresponding segmentation mask were cropped or padded to $200 \times 200$ to guarantee a uniform input for the model.
Then we obtained the brain mask using simple thresholding and mask filling. Gaussian normalization was applied to each subject to rescale the intensities. Dice loss function \cite{milletari2016v} was employed during the training process. Data augmentation including rotation, shearing and zoom was used during the batch training. The optimal number of epochs was set to 50 by contrasting training loss and validation loss over epochs. The batch size was set to 30 and learning rate was set to 0.0002 throughout all of the experiments.

\section{Materials}
\subsection{Datasets and Experimental Setting}
Two clinical datasets: the public MICCAI WMH dataset with 60 cases from 3 centers and a private MS Lesion dataset with 30 cases collected from a hospital in Munich, were employed in our experiments. For each dataset, the FLAIR and T1 modality of each subject were co-registered. Properties of the data were summarised in Table \ref{table:Table1}.
In the experiments reported in Section \ref{comparison} and \ref{analysis}, five-fold cross-validation setting was used.
Specifically, subject IDs were used to split the public training dataset into training and validation sets. In each split, slices from 16 subjects from each center were pooled into training set, and the slices from the remaining 4 subjects from each center for testing. This procedure was repeated until all of the subjects had been used in testing phase. The Dice score, lesion recall and lesion F1-score of all testing subjects were averaged afterwards.
\begin{table*}[t]
	\vspace{-0.2cm}
	\scriptsize
	\newcommand{\tabincell}[2]{\begin{tabular}{@{}#1@{}}#2\end{tabular}}
	\renewcommand\arraystretch{1}
	\centering
	\caption{Detailed information of \textit{MICCAI WMH Challenge} dataset from three centers and private \textit{Multiple Sclerosis} dataset from a hospital in Munich.}\label{table:Table1}.
	\begin{tabular}{clccccc}
		\toprule
		\textbf{Datasets} &\textbf{~Lesion Type~} &\textbf{~Subjects~}&\textbf{~Voxel Size $(m^3)$} &\textbf{~~Size of FLAIR\&T1 Scans}\\
		\midrule
		{Utrecht} &~~~WMH &~~~20 & 0.96$\times$0.95$\times$3.00 & 240$\times$240$\times$48  \\
		{Singapore}&~~~WMH&~~~20 & 1.00$\times$1.00$\times$3.00 & 252$\times$232$\times$48 \\
		{GE3T}&~~~WMH &~~~20& 0.98$\times$0.98$\times$1.20&132$\times$256$\times$83 \\ \midrule
		{Munich}&~~Multiple Sclerosis& ~~~30& 1.00 $\times$1.00$\times$0.99&240$\times$240$\times$170 \\ \midrule
	\end{tabular}

\end{table*}
	\vspace{-0.2cm}
\subsection{Evaluation Metrics}
Three evaluation metrics were used to evaluate the segmentation performance of the algorithm in different aspects from MICCAI WMH Challenge. Given a ground-truth segmentation map $G$ and a segmentation map $P$ generated by an algorithm, the evaluation metrics are defined as follows. \textbf{Dice score}: \emph{DSC} = {2(G$\cap${P})}/({$|G|+|P|$}). This metric measures the overlapping volume of $G$ and $P$. \textbf{Recall for individual lesions}: Let $N_{G}$ be the number of individual lesions delineated in $G$, and $N_{P}$ be the number of correctly detected lesions after comparing $P$ and $G$. Each individual lesion is defined as a 3D connected component. Then the recall for individual lesions is defined as: \emph{Recall} = ${N_P}/{N_G}$. \textbf{F1-score for individual lesions}: Let $N_{P}$ be the number of correctly detected lesions after comparing $P$ and $G$. $N_{F}$ be the number of wrongly detected lesions in $P$. Each individual lesion is defined as a 3D connected component. Then the recall for individual lesions is defined as: \emph{F1} = ${N_P}$/($N_{P}+N_{F}$).


\section{Results}
\subsection{Comparison with the State-of-the-Art} \label{comparison}
\begin{figure*}[]
	\begin{center}
		\includegraphics[width=0.95 \linewidth,height=0.32\linewidth]{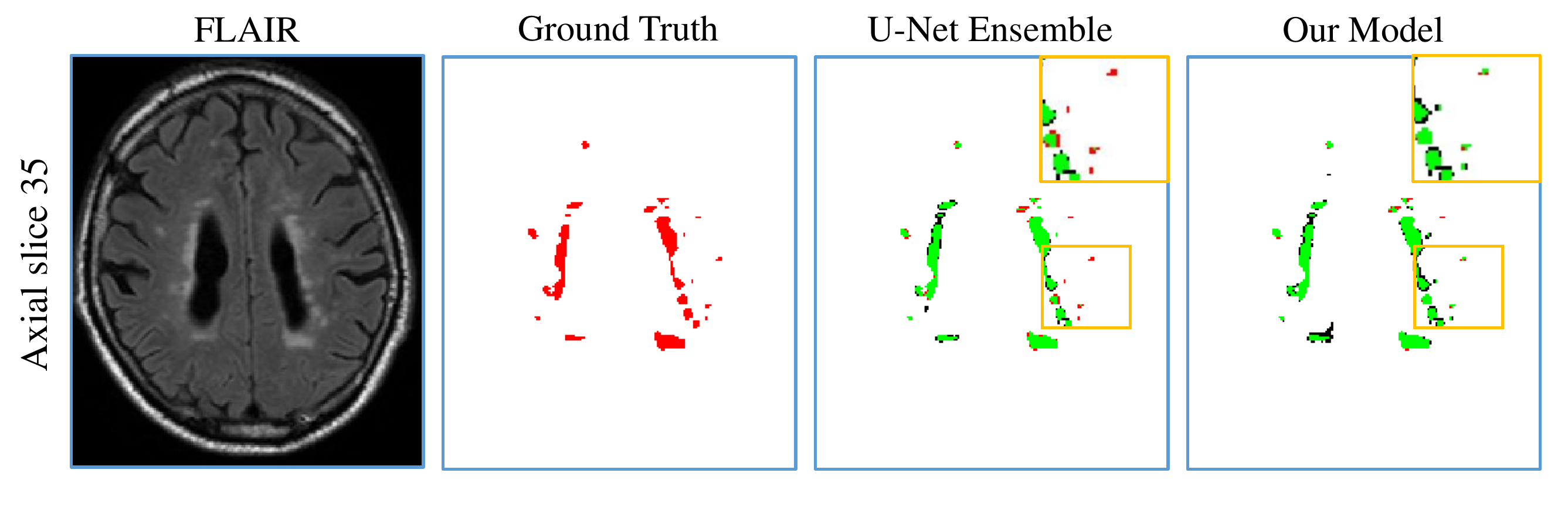}
	\end{center}
	\vspace{-0.2cm}
	\caption{Segmentation result of a testing case from \emph{Utrecht} by U-Net ensembles and our model respectively. The green area is the overlap between the segmentation result and the ground truth. The red pixels are the false negatives, and the black ones are the false positives.}

	\label{fig:segmentationResults}
\end{figure*}

We conducted experiments on the public MICCAI WMH Challenge dataset (3 subsets, 60 subjects) in a 5-fold cross validation setting.
We compared the segmentation performance of the proposed \emph{Stack-Net} and aggregation model with the winning
method in MICCAI WMH Challenge 2017. The \emph{Stack-Net} with 3$\times$3 kernel slightly outperforms \emph{U-Net ensembles} on Dice score and lesion F1-score and achieved comparable lesion recall. The aggregation model outperforms \emph{U-Net ensembles} by 4\% on lesion recall, suggesting that the \emph{Stack-Net} is capable of learning attributes of small-volume lesions. We conducted a paired Z-test over the 60 pairs, where each pair is the lesion recall values obtained on one validation scan by the proposed aggregation model and \emph{U-Net ensembles}. Small p-value (p$<$0.01) indicates that the improvements are statistically significant.
Fig. \ref{fig:segmentationResults} shows a segmentation case in which we can see that our aggregation model is more effective in detecting small lesion.
Furthermore, the proposed method claimed to be the \textbf{first place} (teamname: $sysu\_media\_2$) on the hidden set after independent evaluation by the challenge organizer. Please see the details in {\url{http://wmh.isi.uu.nl/results/}}.

\begin{figure*}[]
	\begin{center}
		\includegraphics[width=0.95 \linewidth,height=0.41\linewidth]{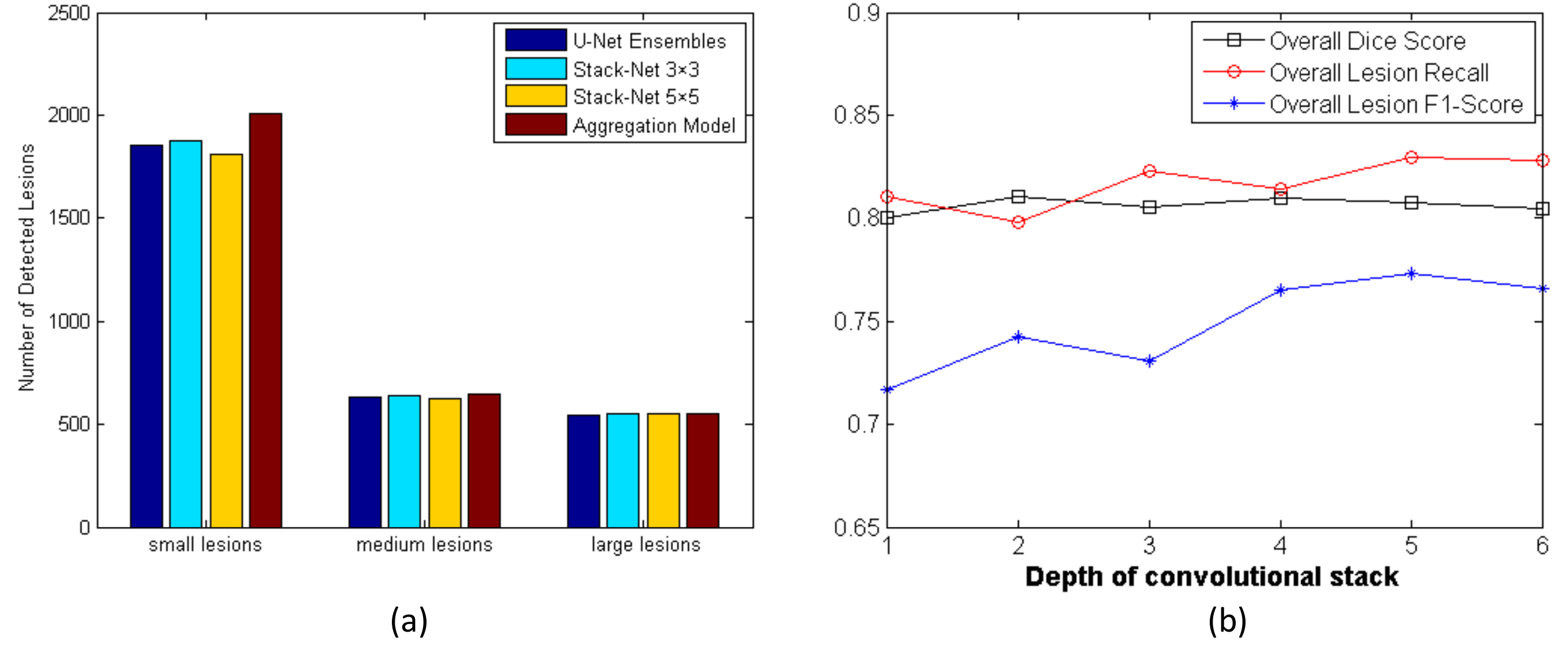}
	\end{center}
	\vspace{-0.1cm}
	\caption{\textbf{a}. Distribution of small, medium and large lesions detected by U-Net ensembles, each component of our aggregation model; \textbf{b}. Overall Dice, lesion recall and lesion F1-score achieved by six \emph{Stack-Nets} with different depths.}

	\label{fig:distribution}
\end{figure*}

To better understand how each part of the proposed model worked effectively on the volume-varied lesions, we grouped all the lesions into three types: small, medium and large, by defining the volume range of each type. Three sets $S_{small} = \{s|~volume(lesion)<10\}$, $S_{medium} = \{s|~10<volume(lesion)<20\}$ and $S_{large} = \{s|~volume(lesion)>20\}$ were obtained.
Then the number of detected lesions of three types was calculated by comparing the predicted segmentation masks and ground-truth segmentation masks on all the test subjects.
Fig. \ref{fig:distribution}(a) further shows the distribution of detected lesions with small, medium and large volumes respectively. Our aggregation model detected 2008 small lesions while the U-Net ensembles detected 1851, i.e., \textbf{8\%} improvement over U-Net ensembles. We conducted a paired Z-test over the 60 pairs, where each pair is the recall values of small lesion obtained on one validation scan by the proposed aggregation model and \emph{U-Net ensembles}. Small p-value (p$<$0.01) indicates that the improvements are statistically significant.
We also observed that the aggregation model achieved a comparable Dice score which measures the overlapping volumes,
demonstrating that it was effective in dealing with both large and small lesions. The \emph{Stack-Net} with 5$\times$5 kernel slightly outperformed U-Net ensembles in the detection of large lesions. It demonstrates that large convolutional kernel is effective in learning contextual information from large abnormality regions with spatial continuity.

\begin{table*}[t]
	\scriptsize
	\newcommand{\tabincell}[2]{\begin{tabular}{@{}#1@{}}#2\end{tabular}}
	\renewcommand\arraystretch{1}
	\centering
	\caption{Comparsion with the winning method in MICCAI WHM Challenge 2017. Values in bold indicates results outperforming the state-of-the-art.}\label{table:TableStateOfTheArt}.
	\begin{tabular}{c | c | c | c }
		\hline
		\textbf{Method}&~~\textbf{Dice Score}~~&~~\textbf{Lesion Recall}~~&~~\textbf{Lesion F1-Score}~~\\
		\hline
		\emph{U-Net ensembles \cite{li2018fully}}& 80.10\% & 82.96\%  & 76.41$\%$\\
        \hline
        \emph{Stack-Net with 3$\times$3 kernel(ours)}& \textbf{80.75}\% & 82.94\%  & \textbf{77.29}$\%$\\
        \emph{Stack-Net with 5$\times$5 kernel(ours)}& \textbf{80.46}\% & 80.96\%  & \textbf{76.75}$\%$\\
		\emph{Multi-scale aggregation model (ours)}& 80.09\% & \textbf{86.96}\% & \textbf{76.73}$\%$\\
        \hline
	\end{tabular}

\end{table*}
 \vspace{-0.1cm}

 \vspace{-0.1cm}
\subsection{Analysis on the \emph{Stack-Net}}  \label{analysis}
To investigate the effect of the depth in \emph{Stack-Net}, we evaluate six models with 3$\times$3 kernel with depths ranging from 1 to 6 on the MICCAI WMH dataset using 5-fold cross validation. Using the same grouping criteria and calculation strategy as mentioned above, we calculate the averaged Dice, averaged lesion recall and averaged lesion F1-score on all test subjects after 5 splits.
As one can observe from Fig. \ref{fig:distribution}(b) that using the \emph{thin} convolutional stack i.e., one or two convolutional layer yields relatively poor segmentation performance on three evaluation metrics.
This is because spatial information is reduced drastically after the sub-sampling layer while the thin stack is not able to preserve rich information and the reduced spatial information is propagated to the deconvolutional layers.
\vspace{-0.2cm}
\subsection{Cross-Center Evaluation on the MS Lesion Dataset}
\begin{table*}[t]

	\scriptsize
	\newcommand{\tabincell}[2]{\begin{tabular}{@{}#1@{}}#2\end{tabular}}
	\renewcommand\arraystretch{1}
	\centering
	\caption{Segmenation performance on the MS lesion dataset. Figures in bold indicate the best performance.}\label{table:MS_lesion}.
	\begin{tabular}{c | c | c | c }
		\hline
		\textbf{Method}&~~\textbf{Dice Score}~~&~~\textbf{Lesion Recall}~~&~~\textbf{Lesion F1-Score}~~\\
        \hline
        \emph{U-Net ensembles \cite{li2018fully}}& 75.95\% & 93.16\%  & 42.41$\%$\\
        \emph{Stack-Net with 3$\times$3 kernel (ours)}& {75.89}\% & \textbf{94.76}\%  & {42.59}$\%$\\
        \emph{Stack-Net with 5$\times$5 kernel (ours)}& {74.71}\% & 93.37\%  & {41.15}$\%$\\
		\emph{Multi-scale aggregation model (ours)}& \textbf{76.93}\% & {93.16}\% & \textbf{49.57}$\%$\\
        \hline
	\end{tabular}
	\vspace{-0.1cm}
\end{table*}

To further evaluate the idea of multi-scale spatial aggregation in a cross-center-evaluation manner, we trained the models on MICCAI WMH dataset, and tested them on the MS lesion dataset from a hospital in Munich.
MS lesions have a very similar appearance with WM lesions, but most of them are medium or large lesions.
Table \ref{table:MS_lesion} reported resulted from a comparison of segmentation performance of individual network and aggregation model.
We observed that the aggregation model achieved significantly better lesion F1-score compared to individual networks, suggesting combination of multi-scale spatial information can help to remove false positives.
Interestingly, we found the lesion recall did not improve after aggregating the individual \emph{Stack-Nets}.
This is due to the fact that most of the MS lesions are in medium or large size, which made the function of convolutional stack achieve limited improvement over the lesion recall. It further suggested that the aggregation of models with multi-scale receptive field is effective in learning multi-scale spatial information.

\section{Conclusions}
In this paper, we explored an architecture specifically designed for small lesion segmentation, to learn attributes of small regions.
We found the convolutional stack was effective in preserving local information of small lesions and the rich information was propagated to the high-resolution deconvolutional stack. By aggregating multi-scale \emph{Stack-Net} with different receptive fields, our method outperformed the state-of-the-art on MICCAI WMH Challenge dataset. We further showed multi-scale context aggregation model was effective in MS lesion segmentation under a cross-center evaluation.

\section{Acknowledgements}
This work was supported in part by NSFC grant (No.~61628212), Royal Society International Exchanges grant (No.~170168),
the Macau Science and Technology Development Fund under 112/2014/A3.
We gratefully acknowledge the support of NVIDIA Corporation with the donation of the Titan Xp GPU used for this research.

{
\bibliographystyle{nature}
\bibliography{egbib}

\begin{thebibliography}{1}
\providecommand{\url}[1]{\texttt{#1}}
\providecommand{\urlprefix}{URL }

\bibitem{borghesani2013association}
Borghesani, P.R., Madhyastha, T.M., Aylward, E.H., Reiter, M.A., Swarny, B.R.,
  Schaie, K.W., Willis, S.L.: The association between higher order abilities,
  processing speed, and age are variably mediated by white matter integrity
  during typical aging. Neuropsychologia  51(8),  1435--1444 (2013)

\bibitem{debette2010clinical}
Debette, S., Markus, H.: The clinical importance of white matter
  hyperintensities on brain magnetic resonance imaging: systematic review and
  meta-analysis. Bmj  341,  c3666 (2010)

\bibitem{li2018fully}
Li, H., Jiang, G., Wang, R., Zhang, J., Wang, Z., Zheng, W.S., Menze, B.: Fully
  convolutional network ensembles for white matter hyperintensities
  segmentation in mr images. arXiv preprint arXiv:1802.05203v1  (2018)

\bibitem{long2015fully}
Long, J., Shelhamer, E., Darrell, T.: Fully convolutional networks for semantic
  segmentation. In: Proceedings of the IEEE conference on computer vision and
  pattern recognition. pp. 3431--3440 (2015)

\bibitem{menze2015multimodal}
Menze, B.H., Jakab, A., Bauer, S., Kalpathy-Cramer, J., Farahani, K., Kirby,
  J., Burren, Y., Porz, N., Slotboom, J., Wiest, R., et~al.: The multimodal
  brain tumor image segmentation benchmark (brats). IEEE transactions on
  medical imaging  34(10),  1993--2024 (2015)

\bibitem{milletari2016v}
Milletari, F., Navab, N., Ahmadi, S.A.: V-net: Fully convolutional neural
  networks for volumetric medical image segmentation. In: 3D Vision (3DV), 2016
  Fourth International Conference on. pp. 565--571. IEEE (2016)

\bibitem{moeskops2018evaluation}
Moeskops, P., de~Bresser, J., Kuijf, H.J., Mendrik, A.M., Biessels, G.J.,
  Pluim, J.P., I{\v{s}}gum, I.: Evaluation of a deep learning approach for the
  segmentation of brain tissues and white matter hyperintensities of presumed
  vascular origin in mri. NeuroImage: Clinical  17,  251--262 (2018)

\bibitem{ronneberger2015u}
Ronneberger, O., Fischer, P., Brox, T.: U-net: Convolutional networks for
  biomedical image segmentation. In: International Conference on Medical image
  computing and computer-assisted intervention. pp. 234--241. Springer (2015)

\bibitem{yu2015multi}
Yu, F., Koltun, V.: Multi-scale context aggregation by dilated convolutions.
  arXiv preprint arXiv:1511.07122  (2015)

\end{thebibliography}
}

\end{document}